\documentclass[preprint,12pt]{elsarticle}

\usepackage{amssymb}
\usepackage{adjustbox}
\usepackage{subcaption}
\usepackage{longtable}
\usepackage{lscape}
\usepackage{pdflscape}
\usepackage{multirow}
\usepackage{multicol}
\usepackage{tabularray}
\usepackage{booktabs}
\usepackage{xcolor}
\usepackage[normalem]{ulem}

\journal{XXXXXXXXXXXX}

\begin{document}

\begin{frontmatter}



\title{Enhancing Generalization in Sickle Cell Disease Diagnosis through Ensemble Methods and Feature Importance Analysis}
%\large{Red Blood Cell Classification}

% \author[inst1]{Author One}

% \affiliation[inst1]{organization={Department One},%Department and Organization
%             addressline={Address One}, 
%             city={City One},
%             postcode={00000}, 
%             state={State One},
%             country={Country One}}

\author[1]{Nataša Petrović}\ead{npe785@uib.es}
\author[1,2]{Gabriel Moyà-Alcover}\ead{gabriel.moya@uib.es}
\author[1,2]{Antoni Jaume-i-Capó\corref{cor1}}\ead{antoni.jaume@uib.es}
\author[1]{Jose Maria Buades Rubio}\ead{josemaria.buades@uib.es}

\cortext[cor1]{Corresponding author}
\affiliation[1]{UGiVIA research group. University of the Balearic Islands. Dpt. of Mathematics and Computer Science. 
Crta. Valldemossa, km 7.5, E-07122 Palma, Spain}

\affiliation[2]{Laboratory for Artificial Intelligence Applications (LAIA@UIB), University of the Balearic Islands, 07122 Palma (Spain)}

\begin{abstract}
This work presents a novel approach for selecting the optimal ensemble-based classification method and features with a primarly focus on achieving generalization, based on the state-of-the-art, to provide diagnostic support for Sickle Cell Disease using peripheral blood smear images of red blood cells. We pre-processed and segmented the microscopic images to ensure the extraction of high-quality features. To ensure the reliability of our proposed system, we conducted an in-depth analysis of interpretability. Leveraging techniques established in the literature, we extracted features from blood cells and employed ensemble machine learning methods to classify their morphology. Furthermore, we have devised a methodology to identify the most critical features for classification, aimed at reducing complexity and training time and enhancing interpretability in opaque models. Lastly, we validated our results using a new dataset, where our model overperformed state-of-the-art models in terms of generalization. The results of classifier ensembled of Random Forest and Extra Trees classifier achieved an harmonic mean of precision and recall (F1-score) of 90.71\% and  a Sickle Cell Disease diagnosis support score (SDS-score) of 93.33\%. These results demonstrate notable enhancement from previous ones with Gradient Boosting classifier (F1-score 87.32\% and SDS-score 89.51\%). To foster scientific progress, we have made available the parameters for each model, the implemented code library, and the confusion matrices with the raw data.

\end{abstract}




\end{frontmatter}


\section{Introduction}
%\subsection{SCD}
Sickle Cell Disease (SCD) is a genetic blood disorder affecting millions of people worldwide. It is caused by a mutation in the haemoglobin gene, which produces abnormal hemoglobin molecules that can distort red blood cells into a crescent or sickle shape. The distorted cells can clog blood vessels, leading to severe pain and organ damage. Although the disease is not curable, it could be prevented and treated. Accurate classification of SCD is essential for effective treatment and management of the disease.

\subsection{\textcolor{red}{Literature Review}}
\textcolor{blue}{Machine learning (ML)} techniques have been increasingly applied to classifying SCD, with promising results \textcolor{blue}{in recent years}. For Red Blood Cell (RBC) classification mostly k Nearest Neighbours (kNN) \cite{lotfi2015detection, rodrigues2016morphological, sharma2016anemia}, \textcolor{blue}{Support Vector Machine (SVM)} \cite{chy2018automatic} and \textcolor{blue}{ Neural Networks (NN)} \cite{tomari2014computer, veluchamy2012feature} were used for both binary and multiclass tasks. \textcolor{blue}{Dasariraju et al.} employed Random Forest (RF)  to classify leukocytes into four classes~\cite{dasariraju2020detection}. 
Ensemble methods have emerged as a valuable approach in various domains, as demonstrated by their effectiveness in multiple applications \cite{zhang2012ensemble}. By combining multiple weaker classifiers,  ensemble methods aim to improve the overall classification performance. 
 This approach \textcolor{blue}{proved} to be effective in different medical diagnosis tasks.  Three ensemble classification models have been employed in \cite{oliveira2017skin}. They were based on the feature selection Optimum-Path Forest classifier and majority voting. Such ensemble methods leverage multiple classifiers' diversity and collective decision-making, leading to enhanced accuracy, robustness, and reliability in classification tasks. Using ensemble methods in medical diagnosis contributes to more accurate and informed decision-making.

 In diabetic retinopathy image classification, Odeh et al. \cite{odeh2021diabetic} ensembled RF, NN and SVM to improve classification accuracy. Five weak classifiers with a weighted vote system were used in \cite{li2020performance}. They outperformed both hard and soft voting systems for cancer classification. With an ensemble of logistic regression, SVM, RF, \textcolor{blue}{eXtreme Gradient Boosting (XGBoost)} and \textcolor{blue}{NN}, they classified 14 types of cancer with 71.46\% accuracy. Another study by Schaefer et al. \cite{schaefer2014ensemble} focused on melanoma diagnosis and developed an ensemble of classifiers, utilizing a one-layer perceptron as a fuser to determine weights for each classifier. Xiong et al. \cite{xiong2021cancer} developed a Naive Bayes Stacking Ensemble, combining SVM, \textcolor{blue}{kNN}, \textcolor{blue}{Decision Trees (DT)}, and  RF classifiers for cancer classification. They demonstrated that this approach generalizes well across nine different binary cancer datasets. Kumar et al. \cite{kumar2022optimized} introduced an optimized model for cancer diagnosis, determining the most effective combination of base classifiers. Maity et al. \cite{maity2017ensemble} successfully demonstrated the efficient classification of red blood cells using a classifier ensemble. These studies collectively showcase the effectiveness and versatility of ensemble methods in improving classification performance across various domains. \textcolor{red}{In fields different than medical image analysis, various authors have demonstrated the usefulness of different \textcolor{blue}{ML} models, ensemble methods and feature importance for modeling complex patterns \cite{jin2024forecasting}, \cite{jin2024price}, \cite{xu2021corn}, \cite{xu2021house}, \cite{zhang2020solubility}, \cite{yu2023remsf},} \textcolor{blue}{\cite{zhang2021estimation}, \cite{zhang2022slope}, \cite{xu2023dynamic}, \cite{phoon2023future}.}

\subsection{\textcolor{red}{Research objective}}
One notable shortcoming of these studies is the omission of an analysis of method performance on datasets that were not a part of the training phase, dealing with diverse capture devices, varying capturing conditions, and differing calibration settings.
%\subsection{objective}
Despite using ensemble approaches in various medical classification tasks, most studies do not employ classifier ensemble for blood cell classification.

%\textcolor{blue}{SCD classification problem is a perfect candidate for ML methods to solve. Recent studies show promising results when using various ML classifiers for blood cell classification. To improve a single classifier's performance and reliability, an ensemble of classifiers emerged as a solution. By combining multiple classifiers, ensemble methods can effectively mitigate the limitations and biases inherent in individual models, leading to improved accuracy and robustness. This approach leverages the diverse strengths of different algorithms, resulting in more comprehensive decision-making processes. Ensembles not only enhance generalization by reducing overfitting but also increase the stability of predictions in varied datasets. 
%Despite using ensemble approaches in various medical classification tasks, most studies do not employ classifier ensembles for blood cell classification. \\
%Enhancing their interpretability is a crucial step in ensuring trustworthy results from ML models Interpretable models rely on the features as a fundamental component. Therefore, determining the importance of features is a vital outcome that allows a better understanding of how the model classifies the cells.
 %}

The primary goal of our study is to employ ensemble methods that incorporate previously validated classifiers from our prior work \cite{petrovic2020sickle}, with a primary emphasis on attaining generalization. There, we developed a methodology for selecting the most effective classification method and features to facilitate diagnostic support using peripheral blood smear images of red blood cells. \textcolor{blue}{Our work significantly contributes to model generalization and interpretability.} \textcolor{red}{The novelty of our proposal lies in demonstrating a high generalization on unknown datasets. This is crucial for the practical applicability and robustness across different data sources and conditions, which is not commonly addressed in many conventional studies.} \textcolor{blue}{To enhance generalization, we devised a robust methodology for selecting and fine-tuning ensemble classifiers, ensuring our models can effectively handle unseen data from diverse sources. We emphasized interpretability by conducting a detailed feature importance analysis, allowing us to identify and prioritize the most critical features for classification. This approach not only aids in understanding the decision-making process of the models but also builds trust with medical professionals who rely on these systems for diagnostic support. The dual focus on generalization and interpretability ensures that our models are both reliable and more comprehensible, making them highly suitable for real-world medical application.}

Firstly, we presented an ensemble of classification methods previously reported in the literature as effective for the morphology classification of blood cells. Secondly, we determined the optimal parameters of each selected ensemble to improve the classification of sickle cells in blood smear images for an SCD case study. Thirdly, we identified the most effective ensemble of classifiers based on the case study. Fourthly, we determined the most important features of the case study. Finally, we validated the results using a new dataset to demonstrate the proposed methodology's generalizability.

Ensuring trustworthy results in \textcolor{blue}{ML} models involves a crucial step of interpretability. Interpretable \textcolor{blue}{ML}models rely on the features as a fundamental component. Therefore, determining the importance of features is a vital outcome that allows a better understanding of how the model classifies the cells.

This work is organized as follows: Section~\ref{sec:method} presents the applied research methodology. Experiments are described in Section ~\ref{sec:experiments}. Section~\ref{sec:results} presents the results and discusses them.  The conclusion and future work are presented in Section~\ref{sec:conclusion}. 

\section{\textcolor{red}{Research methodology}}
\label{sec:method}
\begin{figure}[t]
             \begin{subfigure}{0.32\textwidth}
                \centering
                \includegraphics{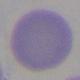}
                \caption{normal}
                \label{fig:normal}
            \end{subfigure}%
            \begin{subfigure}{0.32\textwidth}
                \centering
                \includegraphics{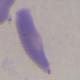}
                \caption{sickle}
                \label{fig:sickle}
            \end{subfigure}%
            \begin{subfigure}{0.32\textwidth}
                \centering
                \includegraphics{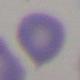}
                \caption{other}
                \label{fig:other}
            \end{subfigure}%
            \centering
            \caption{Examples of the three types of red blood cells to take into account in SCD diagnosis.}
            \label{fig:cell_types}
        \end{figure}

This section presents a methodology for choosing the most effective classification method and features for analyzing cell morphology. Specifically, we focus on supporting SCD diagnosis by considering three types of RBC based on their morphology: normal (discocyte), elongated (sickle cells), and cells with other deformities, as shown in Figure \ref{fig:cell_types}.

\begin{figure}[ht]
\includegraphics[scale=0.4]{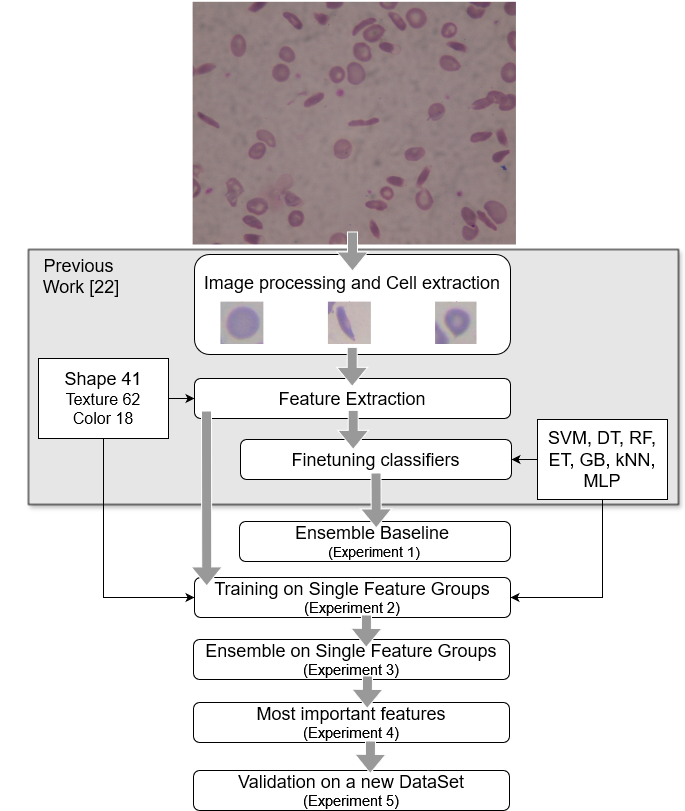}
\centering
\caption{Research methodology outline.}
\label{fig:method}
\end{figure}

\textcolor{red}{The research methodology outline is shown in Figure~\ref{fig:method}. In previous research~\cite{petrovic2020sickle}, we processed a dataset of cell images to isolate individual cells. From these images, we extracted 41 shape features, 62 texture features, and 18 color features. Various classifiers were fine-tuned using these features.}

\textcolor{red}{From these results, we established an ensemble baseline by combining the fine-tuned classifiers' outputs (Experiment 1). Next, we trained classifiers on single feature groups (shape, texture, color) independently to evaluate each group's contribution (Experiment 2). We then created ensembles based on these single feature groups (Experiment 3).}

\textcolor{red}{Feature importance was assessed to identify the most significant features for classification performance (Experiment 4). Finally, we validated the models on a new dataset to ensure their generalization capability and robustness (Experiment 5). }

\textcolor{red}{This methodology systematically evaluates the effectiveness of different classifiers and feature groups in cell image analysis, aiming for accurate and reliable classification. The following subsections describe the feature extraction, classifiers, and ensemble methods used in the research methodology.}

\subsection{Feature extraction}

Feature extraction is the initial step in any classification task using machine learning methods. In the literature on detecting blood cells in microscopic images, several features have been suggested \cite{maity2017ensemble,bhowmick2013structural}. We aimed to distinguish the three types of cells in our classification task.

Scaling images is an essential pre-processing step which helps extract truthful shape features. Before feature extraction, all the images were scaled to the same dimension of individual cells from the dataset in \cite{petrovic2020sickle}. 

Therefore, three types of features were extracted based on the state of the art: shape, color, and texture. To enable the classification of various cell types, we extracted forty-one shape features, eighteen color features, and sixty-two texture features from each cell, resulting in a total of 121 features, as described 
in \cite{petrovic2020sickle}.  \textcolor{red}{Shape features include features such as elongation, aspect ratio, compactness, and various shape factors. Texture features include skewness, kurtosis, and various properties of Gray Level Co-occurrence Matrix (GLCM). Color features include color distribution and intensity features in different color spaces.}

After extraction, all features were standardized, as some classifiers are sensitive to the order of magnitude of the features.

\subsection{Classifiers}
All of the classifiers employed in the experiments were previously documented in the work of Petrovic et al. \cite{petrovic2020sickle}. The selection of these classifiers was guided by their appearance in the scientific literature for the specific task of Red Blood Cell classification.
\subsubsection{Decision trees}

The \textcolor{blue}{DT} classifier is a widely employed \textcolor{blue}{ML} algorithm in various scientific domains. This non-parametric supervised learning method aims to partition the input space into disjoint regions, each associated with a specific class label \cite{hastie2009elements}. DTs are constructed using a top-down approach, where the algorithm recursively splits the feature space based on the selected features and their corresponding thresholds. A decision criterion is applied at each tree node to determine the best feature and threshold combination that optimally divides the data. The resulting tree structure provides an intuitive representation of the decision-making process, as each path from the root to a leaf node corresponds to a specific set of feature conditions and ultimately leads to a predicted class label. One advantage of \textcolor{blue}{DT} classifier is its interpretability and working with numerical and categorical data. However, they are prone to overfitting, mainly when dealing with high-dimensional data or imbalanced class distributions.

\subsubsection{Extra Trees}

The Extra Trees (ET) classifier is an extension of the \textcolor{blue}{DT} classifier, further enhancing the predictive power and generalization capabilities \cite{geurts2006extremely}. Similar to DT, ET constructs an ensemble of trees by recursively partitioning the feature space. However, unlike regular DT, ET introduces additional randomness during training. Specifically, instead of searching for the best feature and threshold at each node, Extra Trees randomly select a set of features and choose random thresholds to split the data. This additional randomization promotes diversity among the trees and helps mitigate overfitting, especially in noisy or high-dimensional datasets. Furthermore, \textcolor{blue}{ETs} offer computational advantages as they can be efficiently parallelized, enabling faster training times for large-scale datasets. The randomness introduced in \textcolor{blue}{ETs} may lead to decreased interpretability compared to traditional Decision Trees.

\subsubsection{Random Forest classifier}

The \textcolor{blue}{RF} classifier is an ensemble learning method that combines multiple \textcolor{blue}{DTs} to enhance prediction accuracy and mitigate overfitting \cite{breiman2001random}. \textcolor{blue}{RFs} leverage the concept of bagging, where each tree in the ensemble is trained on a random subset of the training data, allowing for diversity and reducing the impact of individual noisy or biased samples. RF introduces an additional level of randomness by selecting a random subset of features at each node while constructing individual trees. This process results in diverse trees collectively mak\textcolor{blue}{\sout{e}ing} predictions based on majority voting. \textcolor{blue}{RFs} exhibit several appealing properties, including robustness against overfitting, \textcolor{blue}{handling} high-dimensional data, and \textcolor{blue}{assessing} features' importance for classification. The interpretability of Random Forests may be limited due to the complex interactions and ensemble nature of the model. 

\subsubsection{Gradient Boosting}

The Gradient Boosting (GB) classifier belongs to the family of ensemble learning methods and is renowned for its exceptional predictive performance and versatility. GB operates by iteratively training a sequence of weak learners, typically decision trees, and gradually combining their predictions to form a robust ensemble model \cite{hastie2009elements}. The algorithm aims to minimize a predefined loss function by optimizing the weights assigned to each weak learner based on their performance. During each iteration, the algorithm emphasizes instances previously misclassified, allowing subsequent weak learners to focus on the challenging samples and progressively improve the overall model accuracy. This iterative process, guided by gradient descent, effectively learns complex relationships and interactions within the data, making \textcolor{blue}{GB} particularly adept at handling nonlinear and high-dimensional problems. However, \textcolor{blue}{GB} can be computationally expensive and require careful parameter tuning to prevent overfitting. 

\subsubsection{Suport Vector Machine}
The \textcolor{blue}{SVM} is a supervised learning method that aims to find an optimal hyperplane that separates data points of different classes with the maximum margin \cite{cristianini2000introduction}. The algorithm leverages a kernel function to transform the original input space into a higher-dimensional feature space, where a linear decision boundary can effectively separate the classes. SVM excels in binary classification tasks and can be extended to handle multi-class problems using techniques such as one-vs-one or one-vs-rest. SVM can implicitly capture complex patterns and nonlinear decision boundaries by selecting an appropriate kernel function. Moreover, SVM is known for its robustness against overfitting. SVM's performance may be influenced by the choice of kernel and the selection of appropriate hyperparameters, requiring careful tuning. 

\subsubsection{k-nearest neighbors algorithm}

\textcolor{blue}{kNN} is a non-parametric and instance-based supervised learning method that makes predictions based on the similarity between a query instance and its k nearest neighbors in the training dataset \cite{hastie2009elements}. kNN operates on the assumption that instances with similar feature values tend to belong to the same class. The algorithm determines the class label of a query instance by majority voting among its k nearest neighbors. The choice of k is a crucial parameter in kNN, as a small value may lead to overfitting and high variance, while a large value may introduce bias and oversmoothing. kNN's performance can be affected by higher dimensionality, where the effectiveness of distance-based similarity measures decreases as the feature space grows. 

\subsubsection{Multi Layer Perceptron}

Multi Layer Perceptron (MLP) is a feedforward artificial neural network consisting of multiple layers of interconnected nodes, known as neurons or units \cite{ongun2001feature}. Each neuron in an MLP receives input from the previous layer and applies a nonlinear activation function to produce an output. The algorithm utilizes a training process known as backpropagation, where the network weights are adjusted iteratively to minimize the difference between predicted and actual output values. Careful parameter tuning and architecture selection are essential for achieving optimal results.

\subsection{Ensemble methods}
Ensemble methods are sets of individual classifiers that combine their decisions to achieve more reliable results in machine learning problems \cite{dietterich2000ensemble}. 
 Ensemble classifiers offer significant advantages over individual classifiers in terms of accuracy. 
Constructing an ensemble involves training multiple classifiers on the given dataset, each employing a different learning algorithm or employing the same algorithm with different initializations or subsets of the data. Each classifier independently \textcolor{blue}{ predicts} new examples based on its own learned model. In this study, we used a voting system and stacked generalization (stacking). 
\subsubsection{Voting System}
A voting system of ensemble classifiers refers to a mechanism where multiple individual classifiers collectively contribute to classifying new examples. These individual classifiers form an ensemble, combining their decisions through weighted or unweighted voting  \cite{dietterich2000ensemble}.
The individual classifiers' predictions are combined in the voting system to produce a final classification decision for a new example. Weighted voting assigns different weights to the predictions of individual classifiers based on their reliability or performance. Unweighted voting, on the other hand, treats all individual classifiers equally and counts their predictions equally in the final decision.
By combining the decisions of multiple classifiers, the voting system leverages each classifier's diverse expertise and perspectives. This approach helps mitigate the weaknesses or biases inherent in individual classifiers, improving overall accuracy and robustness. Different classifiers may excel in capturing different aspects of the data or recognizing specific patterns, and the voting system allows for a collective decision-making process that benefits from their combined knowledge.
\subsubsection{Stacked Generalization}

Stacked generalization is a \textcolor{blue}{ML} technique that aims to enhance prediction performance by combining the outputs of multiple individual estimators \cite{wolpert1992stacked}. The process involves two main steps: stacking the predictions of individual estimators and utilizing a classifier to compute the final prediction.
The individual estimators are trained on the given dataset, and their predictions are obtained for each data instance. These estimators can be any type of \textcolor{blue}{ML} models, such as \textcolor{blue}{DT}, \textcolor{blue}{SVM} or \textcolor{blue}{NN}. Each estimator provides its prediction for a given input.
In the stacking phase, the predictions of the individual estimators are combined to create a new feature set. This feature set is constructed by horizontally stacking the predictions, resulting in a matrix where each row corresponds to a data instance and each column represents the prediction from a specific estimator. This newly formed feature set serves as the input for the final estimator.
The final estimator is a classifier that utilizes the stacked predictions as its input to generate the ultimate prediction. This classifier is trained on the stacked feature set and the corresponding ground truth labels. By leveraging the combined knowledge from the individual estimators, the final estimator aims to make more accurate predictions than the individual estimators alone.

\section{Experiments}
\label{sec:experiments}

In this section, we present a detailed account of the five experiments conducted. We evaluated various classifier combinations, examined the feasibility of training on specific feature groups, and employed ensemble techniques to optimize performance. The most effective classifiers were selected, and key features were extracted. Finally, we validated our findings on a new dataset to assess model generalizability.

\subsection{Experiment 1: \textcolor{red}{Evaluating Classifier Combinations}}
% In the first experiment the idea was to test different combinations of classifiers to try to improve the results from previous work. In [citation] is shown that classifier ensemble gives better results. Classifier parameters were taken from [] as finetuned for best possible results on all features. Each combination of classifiers, starting with 2 and progressing to all 7 of them, was run on all features. The result was first decided by voting and then by stacking. Voting classifier is a machine learning estimator that trains various base models or estimators and predicts on the basis of aggregating the findings of each base estimator [citation].
% Stacked generalization consists in stacking the output of an individual estimator and using a classifier to compute the final prediction [citation]. As a final classifier, a Logistic Regression classifier was used.
The initial experiment aimed to evaluate various combinations of classifiers in an attempt to enhance the obtained results. The classifier parameters were obtained from \cite{petrovic2020sickle} and fine-tuned to achieve optimal performance across all features. To comprehensively explore the performance of various classifier combinations, we conducted experiments involving the integration of different classifiers, ranging from 2 to 7, applied to the complete set of features. \textcolor{red}{Classifiers included in the combinations were DT, ET, GB, RF, SVM, kNN and MLP.} The evaluation process involved two independent stages. \textcolor{red}{ In the first stage, we determined the results using a voting mechanism, where each classifier contributed to the final decision based on majority voting. This approach allowed us to assess the collective decision-making power of the combined classifiers. In the second stage, we employed a stacking approach, where the outputs of individual classifiers served as inputs to a meta-classifier, designed to learn the optimal combination of predictions. This method aimed to leverage the strengths of each classifier to achieve superior overall performance.
This detailed analysis provided insights into how different classifiers and their combinations could enhance the classification results, guiding us toward the most promising ensemble strategies.} \textcolor{red}{With this experiment, we were able to calculate the baseline for the ensemble methods using different strategies.}

\subsection{Experiment 2: \textcolor{red}{Training on Single Feature Groups}}
The second experiment aimed to investigate the feasibility of achieving satisfactory results by training a classifier exclusively on a single group of features. Each classifier was trained individually on every feature group (shape, texture and color), allowing for a \textcolor{red}{comprehensive} assessment of the relative importance of each feature type. The primary objective was to ascertain whether one feature type exhibited greater relevance than others in the classification task. 
 By systematically evaluating the performance of classifiers trained on specific feature groups, we sought to gain insights into each feature type's discriminate power and significance. \textcolor{red}{ This analysis enabled us to determine which feature group provided the most critical information for the classification of RBCs, thereby informing future feature selection strategies and model development efforts.}

\subsection{Experiment 3\textcolor{red}{: Ensemble on Specific Features }}
Having trained and fine-tuned classifiers on different feature groups, we decided to ensemble them to improve the results. Each \textcolor{blue}{ classifier was trained  exclusively} on one group of features \textcolor{blue}{ (shape, texture, or color)} and \textcolor{blue}{ combinations were tested to evaluate their collective performance}.  

First, \textcolor{blue}{an} ensemble of 3 classifiers was tested, with the first classifier trained on shape features, the second on texture, and the third on color features. \textcolor{red}{This approach aimed to leverage the unique strengths of each feature type by combining their individual predictions. For this ensemble, we implemented both a voting mechanism, where the final decision was based on \textcolor{blue}{the} majority voting, and a stacking generalization, where a meta-classifier was trained to optimally combine the outputs of the three base classifiers.} 

Secondly, \textcolor{red}{we tested an ensemble of 2 classifiers }where one classifier was trained on shape and the other on texture features. \textcolor{red} {This configuration was chosen to assess whether combining the most informative feature types (as determined by previous analysis) could yield improved results. Similar to the three-classifier ensemble, we used both voting and stacking generalizations to combine the predictions.}

\textcolor{red} {The voting mechanism allowed us to understand the consensus-based performance, while the stacking approach provided insights into the potential for a meta-classifier to enhance prediction accuracy by learning from the base classifiers' outputs.}

\textcolor{red} {These experiments aimed to determine the optimal feature combinations and ensemble methods for RBC classification, providing a robust framework for leveraging multiple feature types and classifier strengths to achieve superior classification performance.}

\subsection{Experiment 4\textcolor{red}{: Most Important Features }}
\textcolor{red}{ The best performing classifiers were selected based on the results of the previous experiment, where various combinations of classifiers and feature groups were tested. In  previous work~\cite{petrovic2020sickle}, we tried different feature selection techniques, such as Principal Component Analysis (PCA), Linear Discriminant Analysis (LDA) and Feature importance. From those results we concluded that the best performing technique was feature importance. We then conducted a detailed feature importance analysis to identify the most critical features for RBC classification within each feature group (shape, texture and color). This analysis utilized techniques such as feature ranking and importance scores from tree-based methods, allowing us to pinpoint the features that contributed most significantly to the classification accuracy.}

\textcolor{red}{Once the most important features were identified, we retrained the selected classifiers using only these key features. By focusing on the most relevant features, we aimed to reduce the model complexity and training time, while maintaining or improving the classification performance.}

\textcolor{red}{After retraining, the classifiers were evaluated using the same performance metrics as in previous experiments to compare the results with those obtained using the full feature sets. This evaluation was crucial to verify that the reduction in feature set did not compromise the model's effectiveness.}

\textcolor{red}{By retraining the classifiers on only the most important features, we aimed to create a more efficient and interpretable model without sacrificing performance, ultimately enhancing the practicality and applicability of our classification system for RBCs.}

\subsection{Experiment 5\textcolor{red}{: Validation on New Dataset}}
Finally, \textcolor{blue}{the fifth experiment, the most important,} includes validating the results on the new dataset to test how well the models generalize. \textcolor{red}{Initially, the original best single classifier (identified in the previous work \cite{petrovic2020sickle}) was run on the new dataset. This step provided a baseline reference for comparison, allowing us to understand how well the standalone classifier performed on unseen data.
The best performing ensemble classifiers from previous experiments was used for validation. Subsequently, the ensemble classifiers were run on the validation dataset. These ensemble classifiers were chosen based on their previous performance and their complementary strengths in handling complex classification tasks. The results from the ensemble classifiers were compared against the reference results obtained from the original classifier. Performance metrics were calculated for both the original and ensemble classifiers to evaluate their effectiveness on the new dataset. The primary focus was to determine if the ensemble classifiers maintained or improved their performance on the unseen dataset compared to the reference classifier. By systematically evaluating the ensemble classifiers on the new dataset, we aimed to validate their generalization capability. This step was crucial to demonstrate that the proposed ensemble method not only performed well on the initial dataset but also reliably handled diverse and unseen data, confirming its robustness and applicability in real-world scenarios.}

\subsection{Experimentation framework and evaluation metrics}
The experimentation framework and evaluation metrics employed in this study were adapted from a previous research \cite{petrovic2020sickle}. 
In this study, two different datasets were used. The first one is ErythrocitesIDB from \cite{petrovic2020sickle} used in first four experiments. The second one, used as a validation dataset, consists of 1440 individual cells. 1099 of those are circular, 192 elongated and 149 labeled as other. The second dataset can be found here: \\ http://erythrocytesidb.uib.es/SCDEnsemble.zip. As in previous experiments, we dealt with an imbalanced dataset, as 76\% of cells are labeled circular.

The chosen metrics, the F1-score and SDS score, were identified as the most pertinent measures for assessing the classification performance. SDS-score \cite{delgado2020diagnosis} is justified by the fact that the misclassification of the normal cells as elongated or other cells will cause the alert the medical specialist that there is a change in the patients' condition.

We employed the same set of classifiers used in the aforementioned paper, ensuring consistency and comparability in evaluating the classification performance across different experiments. By utilizing these established metrics and maintaining uniformity in the classifier selection, we aimed to facilitate direct comparison and evaluation of the classification outcomes in the context of the specific classification task.

\textcolor{red}{All the code was run on 2 GHz Quad-Core Intel Core i5 processor with 16GB RAM.}

\section{Results and Discussion}
%First hypothesis is that color features don’t affect the classification as the previous paper showed that most important features were those of shape and texture. Our second hypothesis states that by ensembling the classifiers, we can improve the results. The final hypothesis is that new method is good in generalization, which it can be showed using it on new dataset.
\label{sec:results}

In this section, results and discussion are presented.
For \textcolor{blue}{the} first  four experiments, ErythrocitesIDB was used as in \cite{petrovic2020sickle}.
As the validation dataset in the fifth experiment SCDEnsemble was used \\ (http://erythrocytesidb.uib.es/SCDEnsemble.zip).
Tables \ref{table:summary1} and \ref{table:summary2} show summary of results from previous work \cite{petrovic2020sickle} for 2 and 3-class classification problem where \textcolor{blue}{the} most performant classifiers were GB and RF.
\begin{center}
\begin{table}[ht!]
\centering
\begin{tabular}{@{}ccc@{}}
\toprule
   & GB   &   RF     
    \\ \midrule
SDS-score & \textbf{95.18\%} & 95.05\%  
\\
F-measure  & \textbf{93.50\%} & 93.36\%  
\\
CBA & \textbf{88.39\%} & 88.06\% 
\\
MCC & \textbf{88.43\%} & 88.20\% 
\\ \hline

\end{tabular}

\caption{Comparison of F-measure and SDS-score values obtained by GB, RF methods for the 2-class classification problem.}
\label{table:summary1}
\end{table}
\end{center}

\begin{center}
\begin{table}[ht!]
\centering
\begin{tabular}{@{}lcccc@{}}
\toprule
    & GB   & RF
    \\ \midrule
SDS-score & \textbf{94.68\%}  & 94.44\%     
\\
F-measure & \textbf{94.67\%} & 94.42\%   
\\
CBA & \textbf{93.98\%} & 93.66\%
\\
MCC & \textbf{88.72\%} & 88.19\%
\\
\hline

\end{tabular}

\caption{Comparison of F-measure and SDS-score values obtained by GB and RF methods  for the 3-class classification problem.} \label{table:summary2}
\end{table}
\end{center}
\subsection{Experiment 1: \textcolor{red}{Evaluating Classifier Combinations}}

In the initial experiment, we conducted an ensemble approach by combining varying numbers of fine-tuned classifiers using all available features. This experimental design aimed to explore the impact of ensemble size on the overall classification performance. By systematically adjusting the number of classifiers included in the ensemble, we sought to investigate any potential relationship between ensemble size and predictive accuracy.  Table \ref{table:ex_1} shows best results obtained with the voting system (see~\ref{appendix-first1} for confusion matrices with raw data).
The results show that the best performing ensembles are MLP and RF, with \textcolor{blue}{the} highest F1 and SDS scores. 

\begin{table}[ht!]
\centering
\begin{tabular}{clcc}
\hline
             \# classifiers & Classifier combination    & F1-score    & SDS-score       \\ \hline
 2 & \textbf{MLP, RF} & \textbf{92.20\%} & \textbf{93.82\%}  \\
 3 & ET, MLP, GB & 91.89\% & 93.70\%  \\
 3 & DT, SVM, MLP & 91.84\% & 93.70\%  \\
3 & SVM, RF, kNN & 91.42\% & 93.70\%  \\
 4 & DT, SVM, MLP, RF & 91.66\% & 93.70\%\\
 4 & DT, RF, GB, kNN & 91.71\% & 93.70\%\\
 5 & ET, SVM, MLP, GB, kNN & 91.35\% & 93.70\% \\
 5 & DT, MLP, RF, GB, kNN & 91.54\% & 93.70\%\\
 5 & ET, DT, SVM, MLP, GB & 91.78\% & 93.70\%\\
 5 & ET, DT, SVM, MLP, kNN & 91.86\% & 93.70\%\\
 \textbf{6} & \textbf{DT, SVM, MLP, RF, GB, kNN} & \textbf{91.72\%} & \textbf{93.82\%}\\
 7 & ET, DT, SVM, MLP, RF, GB, kNN & 91.41\% & 93.33\%\\

\hline
\end{tabular}
\caption{Performance of the different classifiers for the F-measure and SDS-score performance measures in Experiment 1. We show the performance measures in case of voting system.}
\label{table:ex_1}
\end{table}

Table \ref{table:ex_1_stacking} shows best results obtained with the stacking system (see~\ref{appendix-first2} for confusion matrices with raw data).
We can see better results obtained by stacking classifiers rather than \textcolor{blue}{by using} the voting system. The best results were obtained by combining three classifiers DT, SVM and kNN. However, the best combination failed to outperform the single classifier from the last paper, where GB resulted in 93.50\% F1-score and 95.18\% SDS-score and RF in 93.36\% F1-score and 95.05\% SDS-score. 

\begin{table}[ht!]
\centering
\begin{tabular}{clcc}
\hline
             \#  classifiers & Classifier combination    & F1-score    & SDS-score       \\ \hline
 2 & SVM, RF & 91.55\% & 93.82\%  \\
 2 & SVM, kNN & 91.43\% & 93.82\%  \\
 3 & \textbf{DT, SVM, kNN} & \textbf{92.27\%} & \textbf{94.19\%}  \\
 4 & DT, MLP, RF, GB & 92.15\% & 94.07\%\\
 5 & DT, SVM, MLP, GB, kNN & 92.18\% & 94.07\%\\
 6 & DT, SVM, MLP, RF, GB, kNN & 92.01\% & 93.82\%\\
 7 & ET, DT, SVM, MLP, RF, GB, kNN & 91.54\% & 93.45\%\\

\hline
\end{tabular}
\caption{Performance of the different classifiers for the F-measure and SDS-score performance measures in Experiment 1. We show the performance measures in case of stacking system.}
\label{table:ex_1_stacking}
\end{table}

\subsection{Experiment 2: \textcolor{red}{Training on Single Feature Groups}}
 
The best results on different feature groups are shown in Table \ref{table:ex_2} (see~\ref{appendix-second} for confusion matrices with raw data). 
The highest performing classifier was observed to be the Gradient Boosting (GB) algorithm trained on shape features. Conversely, the classifier trained solely on color features exhibited the poorest performance. These outcomes lead us to hypothesize that color features may not be relevant for the classification task.

\begin{table}[ht!]
\centering
\begin{tabular}{lccc}
\hline
              Feature group   & Classifier    & F1-score    & SDS-score       \\ \hline
Shape  & \textbf{GB} & \textbf{92.75\%} & \textbf{94.31\%}  \\
Texture & MLP & 86.49\% & 90.48\%  \\
Color & RF & 77.70\% & 82.32\%\\
\hline
\end{tabular}
\caption{Performance of the different classifiers for the F-measure and SDS-score performance measures in Experiment 2. We show the performance measures in case of training classifiers only with one group of features.}
\label{table:ex_2}
\end{table}

 Our analysis reveals that the shape and texture features alone are sufficient for achieving good performance in the classification task. It is worth noting that images may be captured using different microscope settings, leading to variations in color appearance. Consequently, color properties \textcolor{blue}{are} less reliable \textcolor{blue}{than} features for cell classification. The inherent variability in color across images introduces additional challenges and may hinder the classifier's ability to discern meaningful patterns or characteristics. Therefore, prioritizing shape and texture features, less susceptible to variations in microscope settings, ensures more robust and consistent classification outcomes.

\subsection{Experiment 3\textcolor{red}{: Ensemble on Specific Features }}

The objective of the third experiment was to investigate the effectiveness of ensemble methods by combining multiple classifiers and analyzing their performance. Each classifier was assigned a distinct feature group, with one classifier trained on shape features, another on texture features, and the remaining one on color features. The classifiers were then integrated using a voting system to determine the final prediction. To optimize the ensemble, the parameters of the classifiers were selected based on the best results obtained from a previous experiment. This experimental design allowed for a comprehensive evaluation of ensemble methods in the context of the given classification task, providing insights into the impact of feature groups and parameter selection on the overall performance of the ensemble.

\begin{table}[ht!]
\centering
\begin{tabular}{@{}lccc@{}}
\hline
Classifiers   & F1-score   & SDS-score \\ \hline
$GB_{\textnormal{shape}}$, $MLP_{\textnormal{texture}}$, $RF_{\textnormal{color}}$ & 91.25\% & 93.57\%  \\
$\mathbf{GB_{\textnormal{shape}}, ET_{\textnormal{texture}}, RF_{\textnormal{color}}}$ & \textbf{92.13\%} & \textbf{94.07\%}   \\    
$GB_{\textnormal{shape}}$, $RF_{\textnormal{texture}}$, $RF_{\textnormal{color}}$ & 91.81\% & 93.94\%  \\
$GB_{\textnormal{shape}}$, $RF_{\textnormal{texture}}$, $ET_{\textnormal{color}}$ & 91.79\% & 93.82\% \\
\hline
\end{tabular}

\caption{Comparison of F-measure and SDS-score values obtained by 3 combined classifiers using voting.} \label{table:ex_3}
\end{table}
 
Each classifier used shape, texture and color as features, respectively.
From the obtained results shown in Table \ref{table:ex_3} (see~\ref{appendix-third1} for confusion matrices with raw data), we can observe that the three best combinations are GB, ET and RF with \textcolor{blue}{a} SDS score of 94.07\%, GB, RF and RF with 93.94\% and GB, RF and ET with 93.82\%. Although GB, MLP and RF gave the best results when trained on distinct features combined, they do not outperform other model ensembles.

\begin{table}[ht!]
\centering
\begin{tabular}{@{}lccc@{}}
\hline
     Classifiers   & F1-score   & SDS-score 
    \\  \hline

$\mathbf{GB_{\textnormal{shape}}, GB_{\textnormal{texture}}} $ & \textbf{92.44\%}  & \textbf{94.07\%}        
\\
$GB_{\textnormal{shape}}, ET_{\textnormal{texture}} $ & 92.32\% & \textbf{94.07\%}
\\
$GB_{\textnormal{shape}}, SVM_{\textnormal{texture}} $ & 91.37\% & 93.82\% 
\\
\hline

\end{tabular}

\caption{Comparison of F-measure and SDS-score values obtained by 2 combined classifiers using voting.} \label{table:ex_3_two_voting}
\end{table}

The first part of the experiment included a combination of only two classifiers with only shape and texture features to test our findings that color features are irrelevant for sickle cell classification. The results in Table \ref{table:ex_3_two_voting} (see~\ref{appendix-third2} for confusion matrices with raw data) show that the best combinations are  GB \& GB and GB and ET with 94.07\% of SDS score. This shows that the same results are obtained without one whole group of features. 

The second part of the experiment was designed to ensemble several classifiers and stack their results to obtain the final result shown in Table \ref{table:ex_3_stacking} (see~\ref{appendix-third3} for confusion matrices with raw data). As in the previous one, all groups of features were used for the combination of 3 classifiers; for the other combination, color features were excluded.

\begin{table}[ht!]
\centering
\begin{tabular}{@{}lccc@{}}
\hline
     Classifiers   & F1-score   & SDS-score 
    \\  \hline
    $\mathbf{RF_{\textnormal{shape}}, RF_{\textnormal{texture}}, kNN_{\textnormal{color}}} $ 
 & \textbf{92.98\%}  & \textbf{94.81\% }      
\\
$RF_{\textnormal{shape}}, DT_{\textnormal{texture}}, kNN_{\textnormal{color}} $ & 93.06\% & 94.93\%  
\\
$GB_{\textnormal{shape}}, RF_{\textnormal{texture}}, kNN_{\textnormal{color}} $ & 92.28\% & 94.68\% 
\\

$RF_{\textnormal{shape}}, MLP_{\textnormal{texture}}, kNN_{\textnormal{color}} $ & 92.77\% & 94.68\% \\ \hline

\end{tabular}

\caption{Comparison of F-measure and SDS-score values obtained by 3 ensembled classifiers with stacking.} 
\label{table:ex_3_stacking}
\end{table}

Table \ref{table:ex_3_stacking} shows that the SDS and F1 scores are higher than with voting ensembles. Specifically, RF, DT and kNN with 94.93\%, RF, RF and kNN with 94.81\%. As for 2 classifier combinations with stacking (see Table \ref{table:two_class_stacking} and \ref{appendix-third4} for confusion matrices with raw data), results are even improved in \textcolor{blue}{the} case of  RF, ET with 95.30\% and RF, RF with 95.18\% in the SDS-Score metric.

\begin{table}[ht!]
\centering
\begin{tabular}{@{}lccc@{}}
 \hline
     Classifiers   & F1-score   & SDS-score 
    \\  \hline
     $\mathbf{RF_{\textnormal{shape}}, ET_{\textnormal{texture}}} $ 
  & \textbf{93.35\%}  & \textbf{95.30\%} 
\\
$ RF_{\textnormal{shape}}, RF_{\textnormal{texture}} $ & 93.14\% & 95.18\%  
\\
$ RF_{\textnormal{shape}}, SVM_{\textnormal{texture}} $ & 92.73\% & 94.93\% 
\\ \hline

\end{tabular}

\caption{Comparison of F-measure and SDS-score values obtained by 2 ensembled classifiers with stacking.} \label{table:two_class_stacking}
\end{table}

\subsection{Experiment 4\textcolor{red}{: Most Important Features }}

In this experiment, we extracted \textcolor{blue}{the} most important features from stacked classifiers that obtained the best results, which are given in the following Table \ref{table:important_features}. The best performing classifiers are both interpretable, therefore extraction of \textcolor{blue}{the} most important features was possible.

\begin{table}[ht!]
\centering
\begin{tabular}{@{}cc@{}}
\hline
     Shape   & Texture  
    \\  \hline
elongation & skewness
\\
aspect ratio & kurtosis 
\\
r factor& dissimilarity1
\\
hu1 & dissimilarity3 \\
hu2 & contrast1 \\
fd1 & contrast3 \\
minor axis & contrast5 \\
hu7 & contrast11 \\
roundness & contrast7 \\
shape & dissimilarity4 \\
circularity & contrast2 \\
sphericity & contrast12 \\
modification ratio & contrast4  \\
min feret & \\
compactness & \\
shape factor & \\
hu3 & \\
major axis & \\
max r & \\
max feret & \\  \hline
\end{tabular}

\caption{Most important features obtained with stacking ensemble.} \label{table:two_classes}
\label{table:important_features}
\end{table}

The results with \textcolor{blue}{the} stacking classifiers trained on \textcolor{blue}{the} most important features are shown in the following Table \ref{table:two_stacking}.
Compared to \textcolor{blue}{the} most important features obtained from previous experiments \cite{petrovic2020sickle}, there are 16 features in common, with the addition of sphericity, modification ratio and shape factor. New results show that solidity is not the most important in the stacking ensemble. As for the texture features, skewness, kurtosis and contrast are still important, while homogeneity, energy and correlation do not appear in most important features list.

\begin{table}[ht!]
\centering
\begin{tabular}{@{}lcc@{}}
\hline
    Classifiers   & F1-score   & SDS-score 
    \\  \hline
 $\mathbf{RF_{\textnormal{shape}}, ET_{\textnormal{texture}}} $  & \textbf{93.35\%}  & \textbf{95.30\%}        
\\
$RF_{\textnormal{shape}}, SVM_{\textnormal{texture}} $ & 93.11\% & 95.30\%  
\\
$RF_{\textnormal{shape}}, RF_{\textnormal{texture}} $  & \textbf{93.47\%} & \textbf{95.30\% }
\\
$RF_{\textnormal{shape}}, GB_{\textnormal{shape}} $ & 93.08\% & 95.18\% 
\\
RF \cite{petrovic2020sickle}  & 93.36\% & 95.05\%
\\
GB \cite{petrovic2020sickle}  & 93.50\% & 95.18\%
\\
\hline

\end{tabular}

\caption{Comparison of F-measure and SDS-score values obtained by 2 ensembled classifiers with stacking training on best features and results obtained from previous research.} \label{table:two_stacking}
\end{table}

\subsection{Experiment 5\textcolor{red}{: Validation on New Dataset}}

To validate ensembled classifiers, we used a new dataset labeled by experts (http://erythrocytesidb.uib.es/SCDEnsemble.zip). First,  cells were classified using the best methods from the last paper, \textcolor{blue}{followed by } an ensemble of two classifiers with stacking. The best performing classifiers were stacked RF and ET classifiers. The results are shown in Table \ref{table:validation} (see~\ref{appendix-fourth} for confusion matrices with raw data).

\begin{table}[ht!]
\centering
\begin{tabular}{@{}lcc@{}}
\hline
     Classifiers   & F1-score   & SDS-score 
    \\  \hline
RF \cite{petrovic2020sickle} & 86.20\%  & 89.72\%        
\\
GB \cite{petrovic2020sickle} & 87.32\% & 89.51\%  \\

$\mathbf{RF_{\textnormal{shape}}, ET_{\textnormal{texture}}} $  & \textbf{90.71\%} & \textbf{93.33}\%  

\\

\hline

\end{tabular}
\caption{Comparison of F-measure and SDS-score values obtained on validation dataset.} 
\label{table:validation}
\end{table}

The results obtained from our analysis demonstrate a notable enhancement in both the F1 and SDS score, surpassing the performance of the previous method. These findings strongly indicate the superiority of the new approach in terms of generalization capability. Generalization assumes great significance for multiple reasons. In practical applications, the classification of new SCD cases is vital, necessitating models that can perform reliably on diverse patient populations not initially included in the training dataset. Variability and heterogeneity of SCD manifestations across individuals emphasize the importance of generalization, as it ensures the model's robustness and applicability in clinical settings. Effective generalization minimizes the potential for misdiagnoses,  enhancing the quality of medical practice and contributing to more favorable patient outcomes.

\begin{figure}[ht]
  \subcaptionbox*{Cell predicted as elongated, but labeled as circular.}[.24\linewidth]{%
    \includegraphics[width=\linewidth,height=3cm]{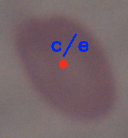}%
  }%
  \hfill
  \subcaptionbox*{Cell predicted as circular, but labeled as other.}[.24\linewidth]{%
    \includegraphics[width=\linewidth,height=3cm]{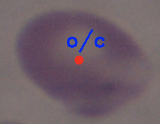}%
  }
  \subcaptionbox*{Cell predicted as other, but labeled as elongated.}[.24\linewidth]{%
    \includegraphics[width=\linewidth,height=3cm]{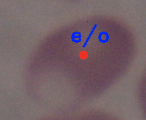}%
  }%
  \hfill
  \subcaptionbox*{Cell predicted as other, but labeled as circular.}[.24\linewidth]{%
    \includegraphics[width=\linewidth,height=3cm]{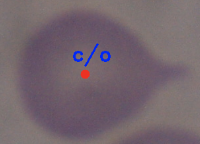}%
  }
  \caption{Circular cells classified as elongated or other.}
\label{fig:cells-1}
\end{figure}

% \begin{figure}[htp]
%     \centering
%     \includegraphics[width=4cm]{prErC.png}
%     \caption{Cell predicted as elongated, but labeled as circular.}
%     \label{fig:cell-1}
% \end{figure}

% \begin{figure}[htp]
%     \centering
%     \includegraphics[width=4cm]{preCrealO.png}
%     \caption{Cell predicted as circular, but labeled as other.}
%     \label{fig:cell-2}
% \end{figure}

% \begin{figure}[htp]
%     \centering
%     \includegraphics[width=4cm]{predictedOrealE.png}
%     \caption{Cell predicted as other, but labeled as elongated.}
%     \label{fig:cell-3}
% \end{figure}

% \begin{figure}[htp]
%     \centering
%     \includegraphics[width=4cm]{predictedOrealC.png}
%     \caption{Cell predicted as other, but labeled as circular.}
%     \label{fig:cell-4}
% \end{figure}

\begin{figure}[ht]
  \subcaptionbox*{Cell predicted as circular, but labeled as other.}[.24\linewidth]{%
    \includegraphics[width=\linewidth,height=3cm]{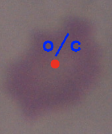}%
  }%
  \hfill
  \subcaptionbox*{Cell predicted as circular, but labeled as elongated.}[.24\linewidth]{%
    \includegraphics[width=\linewidth,height=3cm]{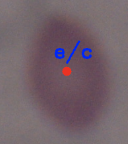}%
  }
  \subcaptionbox*{Cell predicted as circular, but labeled as other.}[.24\linewidth]{%
    \includegraphics[width=\linewidth,height=3cm]{predictedCrealO.png}%
  }%
  \hfill
  \subcaptionbox*{Cell predicted as circular, but labeled as elongated.}[.24\linewidth]{%
    \includegraphics[width=\linewidth,height=3cm]{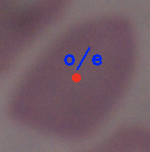}%
  }
  \caption{Elongated or other cells classified as circular.}
 \label{fig: cells-2}
\end{figure}

In Figure \ref{fig:cells-1} are given examples of cells that we consider the predicted class to be the right one, \textcolor{blue}{which} are mislabeled. Given the context of diagnosis, prioritizing false positive examples over false negatives is crucial. Therefore, \textcolor{blue}{when} a circular cell is predicted to be elongated, an expert should be alerted and verify the actual situation rather than overlooking a potentially significant finding. This approach ensures a more cautious and proactive diagnosis, minimizing the risk of missing essential observations or potential abnormalities.
\textcolor{blue}{Figure} \ref{fig: cells-2} \textcolor{blue}{gives} examples of more dangerous misclassification that should be further investigated. False negatives represent a classification error where a test or model inaccurately indicates the absence of a condition or characteristic when it is, in fact, present. False negatives in case of classification of sickle cells are more dangerous than other misclassifications due to their potential to result in delayed treatment and inadequate disease. Analyzing and understanding the reasons behind these more alarming misclassifications are paramount to refining the classifier's performance and enhancing its ability \textcolor{blue}{accurately} to differentiate between different classes, especially minimizing occurrence of false negative results. .

% \begin{figure}[htp]
%     \centering
%     \includegraphics[width=4cm]{predictedCrealO.png}
%     \caption{Cell predicted as circular, but labeled as other}
%     \label{fig:cell-5}
% \end{figure}

% \begin{figure}[htp]
%     \centering
%     \includegraphics[width=4cm]{predCrealE.png}
%     \caption{Cell predicted as circular, but labeled as elongated}
%     \label{fig:cell-6}
% \end{figure}

% \begin{figure}[htp]
%     \centering
%     \includegraphics[width=4cm]{predictedC realO.png}
%     \caption{Cell predicted as other, but labeled as circular}
%     \label{fig:cell-7}
% \end{figure}

% \begin{figure}[htp]
%     \centering
%     \includegraphics[width=4cm]{predErealO.png}
%     \caption{Cell predicted as elongated, but labeled as other}
%     \label{fig:cell-8}
% \end{figure}

Our findings indicate that color features exhibit lower relevance compared to other types of features in the classification task. The newly developed model demonstrates improved generalization performance. Consequently, \textcolor{blue}{including} additional data is expected to have a limited impact. Furthermore, the reduction in the significance of color features suggests that the preprocessing time required for data preparation would be reduced. These insights emphasize the potential benefits of focusing on other feature types while maintaining a robust and efficient classification system.

\textcolor{red}{
\subsection{Pre-processing and classification running times}}
\textcolor{blue}{As we mentioned, All the code was run on 2 GHz Quad-Core Intel Core i5 processor and 16GB RAM.} \textcolor{red}{The table~\ref{table:runnning_times} shows running times for feature extraction, model training and classification. The average running time of extracting all the features of one cell were 1.68s with standard deviation of 0.7s. Without the color features, running time is 1.6s with standard deviation of 0.66s. Using only most important features, running time of feature extraction is down to 1.19s with standard deviation of 0.5s.
Classification of 1440 individual cells with the chosen ensemble takes on average 0.05s with standard deviation of 0.001s. As for training times, the results are as follow:
\begin{itemize}
    \item Training best ensemble of two classifiers with all features $RF_{\textnormal{shape}}$, $ET_{\textnormal{txt}}$ takes on average 9.05s with standard deviation of 0.2s
    \item Training best ensemble of three classifiers with all features $RF_{\textnormal{shape}}$, $RF_{\textnormal{txt}}$, $kNN_{\textnormal{color}}$ takes on average 22.65s with standard deviation of 0.15s.
    \item Training best ensemble of two classifiers with only important features $RF_{\textnormal{shape}}$, $RF_{\textnormal{txt}}$ takes on average 10.31s with standard deviation of 0.18s.
\end{itemize}
}

\begin{table}[h!]
\centering
\begin{tabular}{@{}lcc@{}}
\hline
      \textcolor{red}{\textbf{Event}}  & \textcolor{red}{\textbf{Avg running time}}   & \textcolor{red}{\textbf{Std of running time}} 
    \\  \hline
Feature extraction \\ (all features)& 1.68s  & 0.7s        
\\ \\
Feature extraction \\ (most important features)  & 1.19s & 0.5s  \\
\\
Training best ensemble  \\ (2 classifiers, all features) & 9.05s & 0.2s  \\ \\

Training best ensemble  \\ (3 classifiers, all features)  & 22.65s & 0.15s  \\ \\

Training best ensemble  \\ (2 classifiers, important features)  & 10.31s & 0.18s  \\ \\

Classification on unseen data \\ using best ensemble  & 0.05s & 0.001s  \\

\hline

\end{tabular}
\caption{\textcolor{red}{Feature extraction, model training and classification running times.}} 
\label{table:runnning_times}
\end{table}

\textcolor{blue}{
The proposed model demonstrates high practicability for real-time applications and is cost-efficient based on the observed results. The code was run on a machine whose setup is relatively standard and accessible. The part that takes the longest to execute in the classification pipeline is feature extraction. It would take ~28 minutes to extract features from 1000 cells. The execution time could be decreased by process parallelization. Different machine setups should be considered to further decrease feature extraction runtime. The training running times prove to be very efficient, although model training does not need to be run before every classification task. Previous results indicate that the proposed model is feasible for real-time medical diagnostics and cost-effective, as it requires minimal computational resources and time, making it a viable solution for practical deployment in healthcare settings.}

\section{Conclusion}
Our study focused on the classification of blood cells  using an ensemble of previously reported and effective classification methods. By collecting and implementing features and classifiers validated in the literature for similar cases, we aimed to contribute to the blood cell morphology analysis field. 

Our results include the best parameters for each model, the implemented code available at https://gitlab.com/michonatta/ensemble-classification.git, and confusion matrices with raw data (see \ref{appendix-first}, \ref{appendix-second}, \ref{appendix-third}, and \ref{appendix-fourth} ), facilitating the computation of other metrics by fellow researchers. Furthermore, the dataset we utilized can be accessed at http://erythrocytesidb.uib.es/. To promote scientific progress, we encourage authors to publish their raw data, code, and image datasets used in their studies. Our investigation identified the most effective ensemble of classifiers based on our case study. Additionally, we determined the most important features through feature ranking in the model's prediction output, shedding light on the functioning of opaque models that lack interpretability.

 Ensemble classifier consisted of RF and ET achieved F1-score of 93.35\% and a SDS-score of 95.30\%, while ensemble of two RF classifiers obtained F1 and SDS-scores of 93.47\% and 95.30\%, respectively. We can see that F1-score decreased by 0.01\% and 0.03\% compared to results obtained from individual RF and GB, whereas SDS-score improved by 0.25\% and 0.12\%. 
 
 Moreover, we validated our results using a new dataset \\(http://erythrocytesidb.uib.es/SCDEnsemble.zip), demonstrating the generalizability of our proposed methodology. Notably, our models outperformed state-of-the-art methods in terms of generalization, which is particularly valuable in health environments where trust in diagnostic support systems is crucial. \textcolor{red}{Concretely, the results of our study, with an F1-score of 90.71\% and an SDS-score of 93.33\%, surpass previous state-of-the-art models in terms of generalization (F1-score 86.20\% and 87.32\%, SDS-score 89.72\% and 89.41\%). This significant improvement in performance metrics highlights the effectiveness and novelty of our approach.}

\textcolor{red}{We conducted a detailed feature importance analysis to identify the most critical features for classifying RBCs. This analysis enhances the interpretability of our model while contributing to reduced complexity and training time. Emphasizing interpretability and feature importance in the context of ensemble models adds a unique dimension to our work.}

As a direction for future research, we aim to advance towards more interpretable approaches to enhance the transparency and explainability of our models. Furthermore, we are keen to investigate model generalization further, as adaptability to unseen data holds significant importance for the robustness and effectiveness of the system. \textcolor{red}{Future research can investigate more sophisticated ensemble techniques, such as dynamic ensemble selection or hybrid ensemble methods, to further enhance the model’s performance and adaptability to different types of medical image data. Also, the methodology can be extended to classify other hematological disorders, such as leukemia or malaria. Adapting our approach to different cell types and disease markers can broaden its applicability and benefit a wider range of diagnostic processes. Implanting the proposed method in real diagnostic systems within healthcare facilities will be crucial. Testing the model in real-world clinical environments can provide insights into its practical applicability, user-friendliness, and integration with existing medical software. Finally, although our study emphasizes interpretability, future work can focus on developing more transparent models or using explainable AI (XAI) techniques. Enhancing the interpretability of ensemble methods will help medical professionals understand the decision-making process, thereby increasing trust in automated diagnosis systems.}
\label{sec:conclusion}

\section*{Declaration of Generative AI and AI-assisted technologies in the writing process}
During the preparation of this work the author(s) used ChatGPT in order to to check the correctness of the English in some paragraphs. After using this tool/service, the authors reviewed and edited the content as needed and take full responsibility for the content of the publication.

\section*{Funding}
Project PID2019-104829RA-I00 “EXPLainable Artificial INtelligence systems for health and well-beING (EXPLAINING)” funded by \\ MCIN/AEI/10.13039/501100011033. Project PID2023-149079OB-I00 funded by MICIU/AEI/10.13039/501100011033 and by ERDF/EU.

\section*{Conflict of interest statement}
The authors have declared no conflict of interest.
\raggedright
\bibliographystyle{elsarticle-num} 
\bibliography{bibliography}

\appendix

\section{First experiment: confusion matrices}
\label{appendix-first}
\subsection{Voting}
\label{appendix-first1}

% \begin{longtable}{l}

%\begin{table*}[ht!]
%\resizebox{\textwidth}{!}{%
% \begin{tabular}{@{}lrrrrrrrrrrrrr@{}}
% \cmidrule(r){1-4} \cmidrule(lr){6-9} \cmidrule(l){11-14}
% \begin{tabular}[c]{@{}l@{}}MLP, RF\end{tabular} & c & e & o &  & \begin{tabular}[c]{@{}l@{}}ET, \\MLP, GB \end{tabular} & c & e & o &  & \begin{tabular}[c]{@{}l@{}}DT, \\SVM,  MLP\end{tabular} & c & e & o \\ \cmidrule(r){1-4} \cmidrule(lr){6-9} \cmidrule(l){11-14} 
% c & 473 & 16 & 10 &  & c & 474 & 15 & 10 &  & c & 480 & 13 & 6 \\
% e & 4 & 199 & 7 &  & e & 4 & 200 & 6 &  & e & 5 & 201 & 4 \\
% o & 20 & 6 & 74 &  & o & 22 & 8 & 70 &  & o & 27 & 9 & 64 \\ \cmidrule(r){1-4} \cmidrule(lr){6-9} \cmidrule(l){11-14} 
% \end{tabular}%
%}
%\caption{SVM confusion matrices comparison}
%\label{confusion_svm}
\begin{longtable}{c}
\begin{tabular}{@{}lrrrrrrrrr@{}}
\cmidrule(r){1-4} \cmidrule(lr){5-8} 
\begin{tabular}[c]{@{}l@{}}MLP, RF \end{tabular} & c & e & o &   \begin{tabular}[c]{@{}l@{}}ET, \\MLP, GB \end{tabular} & c & e & o &\\ \cmidrule(r){1-4} \cmidrule(lr){5-8} 
c & 473 & 16 & 10 &   c & 474 & 15 & 10 &    \\
e & 4 & 199 & 7 &   e & 4 & 200 & 6  & \\
o & 20 & 6 & 74 &   o & 22 & 8 & 70 &  \\ \cmidrule(r){1-4} \cmidrule(lr){5-8} 
\end{tabular}%
%}
%\caption{Fine-tuned DT and RF confusion matrices comparison with all features}
\label{confusion_rf_2}
\end{longtable}

\centering
\begin{tabular}{@{}lrrrrrrrrr@{}}

\hline
DT, \\SVM,  MLP  & c & e & o \\
\hline
c & 80 & 13 & 6  \\
e & 5 & 201 & 4   \\
o & 27 & 9 & 64    \\ 
\hline
\end{tabular}

%\resizebox{\textwidth}{!}{%
% \begin{tabular}{@{}lrrrrrrrrrrrrr@{}}
% \cmidrule(r){1-4} \cmidrule(lr){6-9} \cmidrule(l){11-14}
% \begin{tabular}[c]{@{}l@{}}SVM, \\RF, kNN\end{tabular} & c & e & o &  & \begin{tabular}[c]{@{}l@{}}DT, SVM, \\MLP, RF\end{tabular} & c & e & o &  & \begin{tabular}[c]{@{}l@{}}DT, RF, \\GB, kNN\end{tabular} & c & e & o \\ 
% \cmidrule(r){1-4} \cmidrule(lr){6-9} \cmidrule(l){11-14} 
% c & 483 & 13 & 3 &  & c & 478 & 14 & 7 & & c & 476 & 16 & 7  \\
% e & 6 & 197 & 7 &  & e & 5 & 199 & 6 & & e & 5 & 199 & 6 \\
% o & 29 & 9 & 62 &  & o & 25 & 9 & 66 & & o & 23 & 9 & 68 \\ \cmidrule(r){1-4} \cmidrule(lr){6-9} \cmidrule(l){11-14}
% \end{tabular}%
%}
%\caption{Gradient booster confusion matrices comparison}
%\label{confusion_gb}
\begin{longtable}{c}
\begin{tabular}{@{}lrrrrrrrrr@{}}
\cmidrule(r){1-4} \cmidrule(lr){5-8} 
\begin{tabular}[c]{@{}l@{}}SVM, \\RF, kNN \end{tabular} & c & e & o &   \begin{tabular}[c]{@{}l@{}}DT, SVM \\MLP, RF \end{tabular} & c & e & o &\\ \cmidrule(r){1-4} \cmidrule(lr){5-8} 
c & 483 & 13 & 3 &   c & 478 & 14 & 7 &    \\
e & 6 & 197 & 7 &   e & 5 & 199 & 6  & \\
o & 29 & 9 & 62 &   o & 25 & 9 & 66 &  \\ \cmidrule(r){1-4} \cmidrule(lr){5-8} 
\end{tabular}%
%}
%\caption{Fine-tuned DT and RF confusion matrices comparison with all features}
\label{confusion_rf_2}
\end{longtable}

%\centering
\begin{tabular}{@{}lrrrrrrrrr@{}}

\hline
DT, RF, \\GB, kNN  & c & e & o \\
\hline
c & 476 & 16 & 7  \\
e & 5 & 199 & 6   \\
o & 23 & 9 & 68    \\ 
\hline
\end{tabular}

%\resizebox{\textwidth}{!}{%
% \begin{tabular}{@{}lrrrrrrrrrrrrr@{}}
% \cmidrule(r){1-4} \cmidrule(lr){6-9} \cmidrule(l){11-14}
% \begin{tabular}[c]{@{}l@{}}ET, \\SVM, \\MLP, \\GB, kNN\end{tabular} & c & e & o &  & \begin{tabular}[c]{@{}l@{}}DT, \\MLP,  RF,\\ GB, kNN\end{tabular} & c & e & o &  & \begin{tabular}[c]{@{}l@{}}ET, \\DT, SVM, \\MLP, GB\end{tabular} & c & e & o \\ \cmidrule(r){1-4} \cmidrule(lr){6-9} \cmidrule(l){11-14} 
% c & 480 & 14 & 5 &  & c & 478 & 15 & 6 &  & c & 480 & 13 & 6 \\
% e & 5 & 198 & 7 &  & e & 5 & 198 & 7 &  & e & 5 & 198 & 7 \\
% o & 27 & 10 & 63 &  & o & 25 & 9 & 66 &  & o & 27 & 7 & 66 \\ \cmidrule(r){1-4} \cmidrule(lr){6-9} \cmidrule(l){11-14} 
% \end{tabular}%
%}
%\caption{Random forest confusion matrices comparison}
%\label{confusion_rf}
\begin{longtable}{c}
\begin{tabular}{@{}lrrrrrrrrr@{}}
\cmidrule(r){1-4} \cmidrule(lr){5-8} 
\begin{tabular}[c]{@{}l@{}}ET, \\SVM, \\MLP, \\GB, kNN \end{tabular} & c & e & o &   \begin{tabular}[c]{@{}l@{}}DT, \\MLP,  RF,\\ GB, kNN \end{tabular} & c & e & o &\\ \cmidrule(r){1-4} \cmidrule(lr){5-8} 
c & 480 & 14 & 5 &   c & 478 & 15 & 6 &    \\
e & 5 & 198 & 7 &   e & 5 & 198 & 7  & \\
o & 27 & 10 & 63 &   o & 25 & 9 & 66 &  \\ \cmidrule(r){1-4} \cmidrule(lr){5-8} 
\end{tabular}%
%}
%\caption{Fine-tuned DT and RF confusion matrices comparison with all features}
\label{confusion_rf_2}
\end{longtable}

%\centering
\begin{tabular}{@{}lrrrrrrrrr@{}}

\hline
ET, \\DT, SVM, \\MLP, GB  & c & e & o \\
\hline
c & 480 & 13 & 6  \\
e & 5 & 198 & 7   \\
o & 27 & 7 & 66    \\ 
\hline
\end{tabular}
\\

%\resizebox{\textwidth}{!}{%
% \begin{tabular}{@{}lrrrrrrrrrrrrr@{}}
% \cmidrule(r){1-4} \cmidrule(lr){6-9} \cmidrule(l){11-14}
% \begin{tabular}[c]{@{}l@{}}ET, DT, \\SVM, \\MLP, \\kNN\end{tabular} & c & e & o &  & \begin{tabular}[c]{@{}l@{}}DT, SVM,\\MLP, RF, \\GB, kNN\end{tabular} & c & e & o &  & \begin{tabular}[c]{@{}l@{}}ET, \\DT, SVM, \\MLP, RF, \\GB, kNN\end{tabular} & c & e & o \\ \cmidrule(r){1-4} \cmidrule(lr){6-9} \cmidrule(l){11-14} 
% c & 481 & 13 & 5 &  & c & 481 & 13 & 5 &  & c & 477 & 13 & 9 \\
% e & 5 & 199 & 6 &  & e & 5 & 199 & 6  &  & e & 5 & 199 & 6\\
% o & 28 & 7 & 65 &  & o & 27 & 9 & 64 &  & o & 27 & 8 & 65\\ \cmidrule(r){1-4} \cmidrule(lr){6-9} \cmidrule(l){11-14}  
% \end{tabular}%
%}
%\caption{Extra trees confusion matrices comparison}
%\label{confusion_et}

% \end{longtable}

\begin{longtable}{c}
\begin{tabular}{@{}lrrrrrrrrr@{}}
\cmidrule(r){1-4} \cmidrule(lr){5-8} 
\begin{tabular}[c]{@{}l@{}}ET, DT, \\SVM, \\MLP, \\kNN \end{tabular} & c & e & o &   \begin{tabular}[c]{@{}l@{}}DT, SVM,\\MLP, RF, \\GB, kNN \end{tabular} & c & e & o &\\ \cmidrule(r){1-4} \cmidrule(lr){5-8} 
c & 481 & 13 & 5 &   c & 481 & 13 & 5 &    \\
e & 5 & 199 & 6 &   e & 5 & 199 & 6  & \\
o & 28 & 7 & 65 &   o & 27 & 9 & 64 &  \\ \cmidrule(r){1-4} \cmidrule(lr){5-8} 
\end{tabular}%
%}
%\caption{Fine-tuned DT and RF confusion matrices comparison with all features}
\label{confusion_rf_2}
\end{longtable}

%\centering
\begin{tabular}{@{}lrrrrrrrrr@{}}

\hline
ET, \\DT, SVM, \\MLP, RF, \\GB, kNN  & c & e & o \\
\hline
c & 477 & 13 & 9  \\
e & 5 & 199 & 6   \\
o & 27 & 8 & 65    \\ 
\hline
\end{tabular}

\raggedright
\subsection{Stacking}
\label{appendix-first2}
\centering

%\resizebox{\textwidth}{!}{%
\begin{tabular}{@{}lrrrrrrrrrrrrr@{}}
\cmidrule(r){1-4} \cmidrule(lr){6-9} \cmidrule(l){11-14}
\begin{tabular}[c]{@{}l@{}}SVM,RF\end{tabular} & c & e & o &  & \begin{tabular}[c]{@{}l@{}}SVM, kNN\end{tabular} & c & e & o &  & \begin{tabular}[c]{@{}l@{}}DT, \\SVM, kNN\end{tabular} & c & e & o \\ \cmidrule(r){1-4} \cmidrule(lr){6-9} \cmidrule(l){11-14} 
c & 476 & 12 & 11 &  & c & 481 & 11 & 7 &  & c & 486 & 7 & 6 \\
e & 6 & 195 & 9 &  & e & 7 & 194 & 9 &  & e & 8 & 195 & 7  \\
o & 21 & 9 & 70 &  & o & 25 & 9 & 66 &  & o & 26 & 7 & 67 \\ 
\cmidrule(r){1-4} \cmidrule(lr){6-9} \cmidrule(l){11-14} 
\end{tabular}%
%}
%\caption{Decision tree confusion matrices comparison}
%\label{confusion_dt}
%\end{table*}

%\begin{table}[ht!]
%\resizebox{\textwidth}{!}{%
% \begin{tabular}{@{}lrrrrrrrrrrrrr@{}}
% \cmidrule(r){1-4} \cmidrule(lr){6-9} \cmidrule(l){11-14}
% \begin{tabular}[c]{@{}l@{}}DT, \\MLP, \\RF, GB\end{tabular} & c & e & o &  & \begin{tabular}[c]{@{}l@{}}DT, \\SVM, \\MLP, \\GB, kNN\end{tabular} & c & e & o &  & \begin{tabular}[c]{@{}l@{}}DT, SVM, \\MLP, RF, \\GB, kNN\end{tabular} & c & e & o \\ \cmidrule(r){1-4} \cmidrule(lr){6-9} \cmidrule(l){11-14} 
% c & 476 & 12 & 11 &  & c & 480 & 13 & 6  &  & c & 477 & 11 & 11 \\
% e & 5 & 199 & 6 &  & e & 5 & 199 & 6 &  & e & 6 & 198 & 6 \\
% o & 20 & 9 & 71 &  & o & 24 & 8 & 68 &  & o & 22 & 8 & 70 \\
% \cmidrule(r){1-4} \cmidrule(lr){6-9} \cmidrule(l){11-14} 
% \end{tabular}%
%}
%\caption{kNN confusion matrices comparison}
%\label{confusion_knn}

\begin{longtable}{c}
\begin{tabular}{@{}lrrrrrrrrr@{}}
\cmidrule(r){1-4} \cmidrule(lr){5-8} 
\begin{tabular}[c]{@{}l@{}}DT, \\MLP, \\RF, GB \end{tabular} & c & e & o &   \begin{tabular}[c]{@{}l@{}}DT, \\SVM, \\MLP, \\GB, kNN \end{tabular} & c & e & o &\\ \cmidrule(r){1-4} \cmidrule(lr){5-8} 
c & 476 & 12 & 11 &   c & 480 & 13 & 6  &    \\
e & 5 & 199 & 6 &   e & 5 & 199 & 6  & \\
o & 20 & 9 & 71 &   o & 24 & 8 & 68 &  \\ \cmidrule(r){1-4} \cmidrule(lr){5-8} 
\end{tabular}%
%}
%\caption{Fine-tuned DT and RF confusion matrices comparison with all features}
\label{confusion_rf_2}
\end{longtable}

%\centering
\begin{tabular}{@{}lrrrrrrrrr@{}}

\hline
DT, SVM, \\MLP, RF, \\GB, kNN  & c & e & o \\
\hline
c & 477 & 11 & 11   \\
e & 6 & 198 & 6   \\
o & 22 & 8 & 70    \\ 
\hline
\end{tabular}

%\centering
\begin{tabular}{ lccc } 
\hline
ET, DT, \\SVM, \\ MLP, \\RF, GB, \\ kNN & c & e & o \\
\hline
c & 474 & 12 & 13  \\
e & 7 & 197 & 6   \\
o & 21 & 9 & 70    \\ 
\hline
\end{tabular}

\newpage

\raggedright
\section{Second experiment: confusion matrices}
\label{appendix-second}
\centering

% \begin{tabular}{@{}lrrrrrrrrrrrrr@{}}
% \cmidrule(r){1-4} \cmidrule(lr){6-9} \cmidrule(l){11-14}
% \begin{tabular}[c]{@{}l@{}}GB shape \end{tabular} & c & e & o &  & \begin{tabular}[c]{@{}l@{}}MLP texture \end{tabular} & c & e & o &  & \begin{tabular}[c]{@{}l@{}}RF color \end{tabular} & c & e & o \\ \cmidrule(r){1-4} \cmidrule(lr){6-9} \cmidrule(l){11-14} 
% c & 486 & 8 & 5 &  & c & 468 & 16 & 15 &  & c & 461 & 30 & 8 \\
% e & 10 & 198 & 2 &  & e & 56 & 148 & 6 &  & e & 58 & 144 & 8 \\
% o & 23 & 9 & 68 &  & o & 74 & 14 & 12 &  & o & 47 & 19 & 34 \\ \cmidrule(r){1-4} \cmidrule(lr){6-9} \cmidrule(l){11-14} 
% \end{tabular}%

\begin{longtable}{c}
\begin{tabular}{@{}lrrrrrrrrr@{}}
\cmidrule(r){1-4} \cmidrule(lr){5-8} 
\begin{tabular}[c]{@{}l@{}}GB shape \end{tabular} & c & e & o &   \begin{tabular}[c]{@{}l@{}}MLP texture \end{tabular} & c & e & o &\\ \cmidrule(r){1-4} \cmidrule(lr){5-8} 
c & 486 & 8 & 5 &   c & 468 & 16 & 15 &    \\
e & 10 & 198 & 2 &   e & 56 & 148 & 6  & \\
o & 23 & 9 & 68 &   o & 74 & 14 & 12 &  \\ \cmidrule(r){1-4} \cmidrule(lr){5-8} 
\end{tabular}%
%}
%\caption{Fine-tuned DT and RF confusion matrices comparison with all features}
\label{confusion_rf_2}
\end{longtable}

%\centering
\begin{tabular}{@{}lrrrrrrrrr@{}}

\hline
RF color  & c & e & o \\
\hline
c & 461 & 30 & 8  \\
e & 58 & 144 & 8   \\
o & 47 & 19 & 34    \\ 
\hline
\end{tabular}

\newpage

\raggedright
\section{Third experiment: confusion matrices}
\label{appendix-third}
\subsection{3 combined classifiers using voting}
\label{appendix-third1}
\centering

\begin{longtable}{c}

\begin{tabular}{@{}lrrrrrrrrr@{}}
\cmidrule(r){1-4} \cmidrule(lr){5-8} 
\begin{tabular}[l]{@{}l@{}}GB shape, \\MLP txt, \\RF shape\end{tabular} & c & e & o &   \begin{tabular}[c]{@{}l@{}}GB shape,\\ ET txt, \\ RF color\end{tabular} & c & e & o &\\ \cmidrule(r){1-4} \cmidrule(lr){5-8} 
c & 495 & 4 & 0 &   c & 493 & 6 & 0 &    \\
e & 17 & 191 & 2 &   e & 11 & 197 & 2 & \\
o & 31 & 10 & 59 &   o & 31 & 10 & 59 &  \\ \cmidrule(r){1-4} \cmidrule(lr){5-8} 
\end{tabular}%
%}
%\caption{Fine-tuned DT and RF confusion matrices comparison with all features}
\label{confusion_GB_2}

\\

\begin{tabular}{@{}lrrrrrrrrr@{}}
\cmidrule(r){1-4} \cmidrule(lr){5-8} 
\begin{tabular}[l]{@{}l@{}}GB shape, \\RF txt, \\RF color\end{tabular} & c & e & o &   \begin{tabular}[c]{@{}l@{}}GB shape, \\ RF txt, \\ ET color \end{tabular} & c & e & o &\\ \cmidrule(r){1-4} \cmidrule(lr){5-8} 
c & 490 & 8 & 1 &   c & 493 & 6 & 0 &    \\
e & 13 & 196 & 1 &   e & 16 & 193 & 1 & \\
o & 27 & 13 & 60 &   o & 28 & 12 & 60 &  \\ \cmidrule(r){1-4} \cmidrule(lr){5-8} 
\end{tabular}%
%}
%\caption{Fine-tuned DT and RF confusion matrices comparison with all features}
\label{confusion_rf_2}
\end{longtable}

% \begin{tabular}{@{}lrrrrrrrrrrrrr@{}}
% \cmidrule(r){1-4} \cmidrule(lr){6-9} \cmidrule(l){11-14}
% \begin{tabular}[c]{@{}l@{}}GB shape, \\ GB txt \end{tabular} & c & e & o &  & \begin{tabular}[c]{@{}l@{}}GB shape, \\ ET txt \end{tabular} & c & e & o &  & \begin{tabular}[c]{@{}l@{}}GB shape, \\ SVM txt \end{tabular} & c & e & o \\ \cmidrule(r){1-4} \cmidrule(lr){6-9} \cmidrule(l){11-14} 
% c & 493 & 5 & 1 &  & c & 492 & 7 & 0 &  & c & 490 & 8 & 1 \\
% e & 11 & 197 & 2 &  & e & 11 & 197 & 2 &  & e & 15 & 193 & 2 \\
% o & 31 & 8 & 61 &  & o & 30 & 9 & 61 &  & o & 26 & 12 & 62 \\ \cmidrule(r){1-4} \cmidrule(lr){6-9} \cmidrule(l){11-14} 
% \end{tabular}%

\raggedright
\subsection{2 combined classifiers
using voting}
\label{appendix-third2}
\centering

\begin{longtable}{c}

\begin{tabular}{@{}lrrrrrrrrr@{}}
\cmidrule(r){1-4} \cmidrule(lr){5-8} 
\begin{tabular}[c]{@{}l@{}}GB shape, \\ GB txt \end{tabular} & c & e & o &   \begin{tabular}[c]{@{}l@{}}GB shape, \\ ET txt \end{tabular} & c & e & o &\\ \cmidrule(r){1-4} \cmidrule(lr){5-8} 
c & 493 & 5 & 1 &   c & 492 & 7 & 0 &    \\
e & 11 & 197 & 2 &   e & 11 & 197 & 2  & \\
o & 11 & 197 & 2 &   o & 30 & 9 & 61 &  \\ \cmidrule(r){1-4} \cmidrule(lr){5-8} 
\end{tabular}%
%}
%\caption{Fine-tuned DT and RF confusion matrices comparison with all features}
\label{confusion_rf_2}
\end{longtable}

%\centering
\begin{tabular}{@{}lrrrrrrrrr@{}}

\hline
GB shape, \\ SVM txt  & c & e & o \\
\hline
c & 490 & 8 & 1  \\
e & 15 & 193 & 2   \\
o & 26 & 12 & 62    \\ 
\hline
\end{tabular}

%\newpage

\newpage
\raggedright
\subsection{3 combined classifiers
using stacking}
\label{appendix-third3}

\centering

\begin{longtable}{c}

\begin{tabular}{@{}lrrrrrrrrr@{}}
\cmidrule(r){1-4} \cmidrule(lr){5-8} 
\begin{tabular}[c]{@{}l@{}}RF shape, \\ RF txt, \\ kNN shape\end{tabular} & c & e & o &   \begin{tabular}[c]{@{}l@{}}RF shape, \\ DT txt, \\ kNN color\end{tabular} & c & e & o &\\ \cmidrule(r){1-4} \cmidrule(lr){5-8} 
c & 485 & 9 & 5 &   c & 487 & 9 & 3 &    \\
e & 8 & 195 & 7 &   e & 8 & 196 & 6 & \\
o & 20 & 7 & 73 &   o & 21 & 8 & 71 &  \\ \cmidrule(r){1-4} \cmidrule(lr){5-8} 
\end{tabular}%
%}
%\caption{Fine-tuned DT and RF confusion matrices comparison with all features}
\label{confusion_GB_2}

\\

\begin{tabular}{@{}lrrrrrrrrr@{}}
\cmidrule(r){1-4} \cmidrule(lr){5-8} 
\begin{tabular}[c]{@{}l@{}}GB shape, \\ RF txt, \\kNN color\end{tabular} & c & e & o &   \begin{tabular}[c]{@{}l@{}}RF shape, \\ MLP txt, \\ kNN color \end{tabular} & c & e & o &\\ \cmidrule(r){1-4} \cmidrule(lr){5-8} 
c & 488 & 8 & 3 &   c & 487 & 10 & 2 &    \\
e & 11 & 197 & 2 &   e & 8 & 196 & 6 & \\
o & 21 & 8 & 71 &   o & 23 & 8 & 69 &  \\ \cmidrule(r){1-4} \cmidrule(lr){5-8} 
\end{tabular}%
%}
%\caption{Fine-tuned DT and RF confusion matrices comparison with all features}
\label{confusion_rf_2}
\end{longtable}

\raggedright
\subsection{2 combined classifiers
using stacking}
\label{appendix-third4}

\centering

% \begin{tabular}{@{}lrrrrrrrrrrrrr@{}}
% \cmidrule(r){1-4} \cmidrule(lr){6-9} \cmidrule(l){11-14}
% \begin{tabular}[c]{@{}l@{}}RF shape, \\ ET txt \end{tabular} & c & e & o &  & \begin{tabular}[c]{@{}l@{}}RF shape, \\ RF txt \end{tabular} & c & e & o &  & \begin{tabular}[c]{@{}l@{}}RF shape, \\ SVM txt \end{tabular} & c & e & o \\ \cmidrule(r){1-4} \cmidrule(lr){6-9} \cmidrule(l){11-14} 
% c & 487 & 8 & 4 &  & c & 485 & 8 & 6 &  & c & 485 & 9 & 5 \\
% e & 7 & 196 & 7 &  & e & 7 & 195 & 8 &  & e & 8 & 194 & 8 \\
% o & 19 & 8 & 73 &  & o & 18 & 8 & 74 &  & o & 19 & 9 & 72 \\ \cmidrule(r){1-4} \cmidrule(lr){6-9} \cmidrule(l){11-14} 
% \end{tabular}%

%\begin{longtable}
\begin{tabular}{@{}lrrrrrrrrr@{}}
\cmidrule(r){1-4} \cmidrule(lr){5-8} 
\begin{tabular}[c]{@{}l@{}}RF shape, \\ ET txt \end{tabular} & c & e & o &   \begin{tabular}[c]{@{}l@{}}RF shape, \\ RF txt \end{tabular} & c & e & o &\\ \cmidrule(r){1-4} \cmidrule(lr){5-8} 
c & 487 & 8 & 4 &   c & 485 & 8 & 6 &    \\
e & 7 & 196 & 7 &   e & 7 & 195 & 8  & \\
o & 19 & 8 & 73 &   o & 18 & 8 & 74 &  \\ \cmidrule(r){1-4} \cmidrule(lr){5-8} 
\label{confusion_rf_2}
\end{tabular}%
%}
%\caption{Fine-tuned DT and RF confusion matrices comparison with all features}

%\end{longtable}

%\centering
\begin{tabular}{@{}lrrrrrrrrr@{}}

\hline
RF shape, \\ SVM txt  & c & e & o \\
\hline
c & 485 & 9 & 5  \\
e & 8 & 194 & 8   \\
o & 19 & 9 & 72   \\ 
\hline
\end{tabular}

\newpage
\raggedright
\section{Fourth experiment: confusion matrices}
\label{appendix-fourth}
\subsection{2 combined classifiers using stacking on best features}
\centering
\begin{longtable}{c}
\begin{tabular}{@{}lrrrrrrrrr@{}}
\cmidrule(r){1-4} \cmidrule(lr){5-8} 
\begin{tabular}[c]{@{}l@{}}RFshape, ETtexture \end{tabular} & c & e & o &   \begin{tabular}[c]{@{}l@{}} RFshape, SVMtexture \end{tabular} & c & e & o &\\ \cmidrule(r){1-4} \cmidrule(lr){5-8} 
c & 488 & 7 & 4 &   c & 487 & 7 & 5 &    \\
e & 7 & 195 & 8 &   e & 8 & 194 & 8  & \\
o & 20 & 7 & 73 &   o & 18 & 9 & 73 &  \\ \cmidrule(r){1-4} \cmidrule(lr){5-8} 
\end{tabular}%
%}
%\caption{Fine-tuned DT and RF confusion matrices comparison with all features}
\label{confusion_rf_2}
\end{longtable}

\begin{longtable}{c}
\begin{tabular}{@{}lrrrrrrrrr@{}}
\cmidrule(r){1-4} \cmidrule(lr){5-8} 
\begin{tabular}[c]{@{}l@{}}RFshape, RFtexture \end{tabular} & c & e & o &   \begin{tabular}[c]{@{}l@{}}RFshape, GBshape \end{tabular} & c & e & o &\\ \cmidrule(r){1-4} \cmidrule(lr){5-8} 
c & 488 & 8 & 3 &   c & 487 & 9 & 3 &    \\
e & 8 & 195 & 7 &   e & 7 & 195 & 8  & \\
o & 19 & 7 & 74 &   o & 20 & 8 & 72 &  \\ \cmidrule(r){1-4} \cmidrule(lr){5-8} 
\end{tabular}%
%}
%\caption{Fine-tuned DT and RF confusion matrices comparison with all features}
\label{confusion_rf_2}
\end{longtable}

\subsection{Results on validation dataset }
\begin{longtable}{c}
\begin{tabular}{@{}lrrrrrrrrr@{}}
\cmidrule(r){1-4} \cmidrule(lr){5-8} 
\begin{tabular}[c]{@{}l@{}}RF \end{tabular} & c & e & o &   \begin{tabular}[c]{@{}l@{}}GB \end{tabular} & c & e & o &\\ \cmidrule(r){1-4} \cmidrule(lr){5-8} 
c & 1042 & 5 & 52 &   c & 1035 & 4 & 60 &    \\
e & 16 & 153 & 23 &   e & 15 & 167 & 10  & \\
o & 75 & 3 & 71 &   o & 72 & 5 & 72 &  \\ \cmidrule(r){1-4} \cmidrule(lr){5-8} 
\end{tabular}%
%}
%\caption{Fine-tuned DT and RF confusion matrices comparison with all features}
\label{confusion_rf_2}
\end{longtable}

%\centering
\begin{tabular}{@{}lrrrrrrrrr@{}}

\hline
RFshape, \\ ETtxt  & c & e & o \\
\hline
c & 1069 & 4 & 26  \\
e & 0 & 180 & 12   \\
o & 66 & 7 & 76    \\ 
\hline
\end{tabular}





% \end{thebibliography}
\end{document}